\pdfoutput=1
\documentclass[11pt,a4paper]{article}
\usepackage[margin=2.7cm]{geometry}
\usepackage[T1]{fontenc}
\usepackage{lmodern}
\usepackage{microtype}
\usepackage[british]{babel}
\usepackage{graphicx}
\usepackage{xcolor}
\usepackage{tikz}
\usetikzlibrary{positioning,arrows.meta,shapes.geometric,calc}
\usepackage[round,authoryear]{natbib}
\usepackage{enumitem}
\usepackage[hidelinks]{hyperref}

\setlist{itemsep=0.15em,topsep=0.3em}

\title{In Search of Machine Normativity: Towards an Approach to Study Normativity in Large Language Models}
\author{Stavros Anagnou and Christoph Salge\\
\small Adaptive Systems Research Group, University of Hertfordshire, Hatfield, UK\\
\small \texttt{\{s.anagnou,c.salge\}@herts.ac.uk}}
\date{}

\begin{document}
\maketitle

\begin{abstract}
As large language models (LLMs) and other AI systems are integrated into institutions, their individual behaviour no longer explains what they do collectively, whether political bots influence elections or pricing algorithms tacitly collude. Norms are a key ingredient for understanding and steering the emergent behaviour of AI and human--AI collectives, but existing definitions are poorly suited to the task. As \citet{westra2024} argue for animals, prevailing definitions presuppose a human norm psychology, including shared intentionality and advanced belief attribution. Applied to AI, this framing restricts the search to norms produced by human-like cognition and misses norm-like processes that arise through different means. We propose a cognition-agnostic, behavioural definition that looks for norms in group-level regularities rather than individual psychological capacities. Following comparative work, a regularity is normative only if it is socially maintained; we operationalise social maintenance as three measurable properties: minimal social interaction (information flow), a stake, and the means to affect that stake. We add a multiple-equilibria criterion, under which different communities under similar conditions can converge on different stable regularities; this captures the arbitrariness characteristic of norms. The definition yields three tests that can be run on any multi-agent system. We apply them to two LLM case studies and argue that current research either seeks human-like psychology in LLMs or loads stakes into agents explicitly, and so cannot reveal the stakes LLMs bring with them. We propose a two-step approach that first detects normative regularities behaviourally and then investigates the mechanisms behind them, including emulated stakes inherited through training.
\end{abstract}


\section{Introduction}

As LLMs and other AI systems take on more roles in institutions and society, they increasingly interact with each other and with humans. These interactions already have tangible effects: social bots and recommendation algorithms spreading low-credibility content \citep{shao2018,schroeder2026}, trading algorithms implicated in flash crashes \citep{kirilenko2017}, and pricing algorithms learning to collude \citep{calvano2020}. Understanding such collectives requires more than understanding their members individually, because the collective arrangements that can arise cannot be read off individual characteristics alone \citep{epstein1996}.

Norms are among the main ways cooperative human behaviour is organised, so they are a natural lens for this problem. They have also been proposed as an alternative to preferences as the target of AI alignment, so that AI systems would be aligned with the social, legal and moral norms appropriate to their roles rather than with human preferences \citep{zhixuan2024}. Work on the normative capabilities of LLMs is growing. LLMs have been reported to reach conventions without central coordination \citep{ashery2025}, evolve cooperative norms \citep{vallinder2024}, detect norm violations \citep{he2024}, enforce metanorms \citep{horiguchi2024}, and conform to a majority \citep{zhu2025}. Before norms can serve as an alignment target, however, we need a way to study how they arise autonomously among LLMs and to classify them, and there is as yet no agreed framework for doing so.

The obvious candidates come from human norm research, and they disagree about which psychological capacities norms require. Some hold that an intrinsic motivation to follow and enforce norms is essential \citep{sripada2006}; others emphasise shared intentionality \citep{tomasello2020,henrich2016}; others again trace norm psychology to domain-general processes such as reinforcement learning and prediction-error minimisation, or to a culturally evolved ``cognitive gadget'' \citep[reviewed in][]{westra2024}. \citet{westra2024} point out that any definition presupposing a human norm psychology may miss norm-like processes in non-human animals that arise by different cognitive means. They propose instead a cognition-agnostic, behavioural framework that looks for evidence of norms in group-level regularities.

\begin{figure}[t]
\centering
\includegraphics[width=\textwidth]{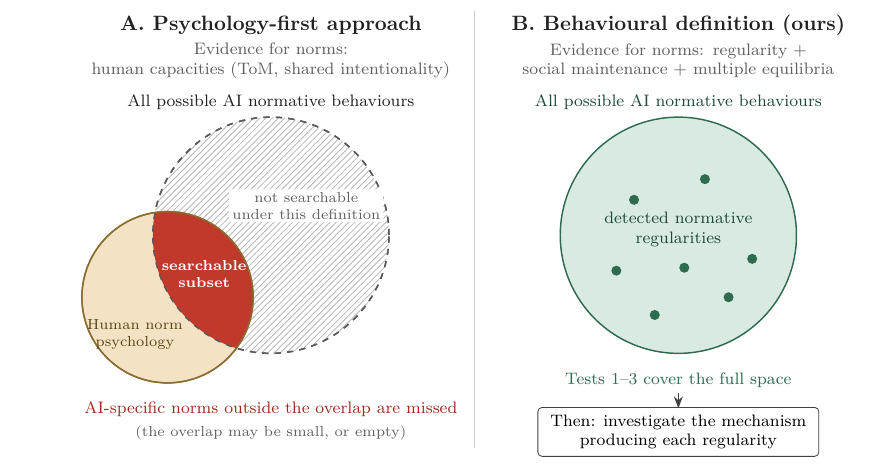}
\caption{Two ways to search for norms in AI collectives. (A) A psychology-first definition treats human normative capacities as the evidence for norms, so only the region where machine and human norm psychology overlap is searchable; AI-specific norms outside it are invisible, and the overlap may be small or empty. (B) A behavioural definition (regularity, social maintenance, multiple equilibria) applies to any agent type and so covers the whole space; once a regularity is detected, the mechanism producing it can be investigated as a separate step.}
\label{fig:space}
\end{figure}

We argue that the same reasoning applies to artefacts such as AI agents. LLMs produce behaviour in a markedly different way from humans, through next-token prediction rather than biologically grounded, value-sensitive judgement \citep{quattrociocchi2025}, and their internals are too complex to map onto behaviour directly. A definition that searches for human capacities in LLMs can only find norms where human and machine norm psychology overlap, and that overlap may be small or empty (Fig.~\ref{fig:space}A). A behavioural definition can search the whole space of normative behaviour and leave the question of mechanism open until a norm has been found (Fig.~\ref{fig:space}B).

This paper makes three contributions. First, we present a behavioural norm definition, \textit{axiomatic normativity}, and derive from it three tests that can be applied to any multi-agent system (Section~\ref{sec:framework}). Second, we review current work on LLM norms and show that it either searches for human capacities or loads stakes into agents explicitly (Section~\ref{sec:review}), and apply our tests to two case studies (Section~\ref{sec:cases}). Third, we outline how, once a norm is detected, the stakes behind it can be investigated, distinguishing stakes loaded through the prompt from emulated stakes inherited through training (Section~\ref{sec:stake}).

\section{Axiomatic normativity}
\label{sec:framework}

\subsection{Normative regularities}

Social norms are the rules of a group that mark out what is appropriate, allowed, required or forbidden for its members in different situations. They typically manifest as behavioural regularities kept in place by social sanctions and social pressure \citep{kelly2025}. Building on \citet{westra2022}, \citet{westra2024} define a \textit{normative regularity} as ``a socially maintained pattern of behavioural conformity within a community''. The pattern of conformity is an observable regularity that supports prediction and generalisation; its existence does not depend on knowing the mechanism that produces it.

Social maintenance is needed because not all conformity is normative. If every animal in a dry landscape drinks at the same water source, that does not make drinking there a norm \citep{westra2024}. The regularity is produced by the environment, a case of \textit{environmental scaffolding}. Similarly, agents that converge on the same behaviour because each independently learns the best response are not socially maintaining it, because no agent is aware of, or affected by, whether another conforms. This too is environmental scaffolding, since each agent finds the best response to a fixed environment.

\subsection{Operationalising social maintenance}

\citet{westra2024} define social maintenance minimally: a pattern is socially maintained ``when individual adherence to the pattern is explained by the way that members of the community react to cases of conformity and non-conformity''. This covers negative social maintenance, which imposes costs on non-conformers (third-party punishment, but also second-party retaliation), and positive social maintenance, which rewards conformity through praise, inclusion, status, or modelling and teaching. To apply this definition to artificial agents we need social maintenance to be measurable. We define it by three properties.

\paragraph{Minimal social interaction.} Agents must be able to affect one another. Following \citet{salge2013}, we treat this information-theoretically: when we can only observe agents, the relevant quantity is the mutual information between their actions; when we can intervene on them, it is the causal information flow from one agent's actions to another's \citep{ay2008}. Information flow has the advantage of measuring the potential for influence, not just influence that happened to occur.

\paragraph{Stake.} For reward or punishment to change behaviour, there must be something the agent cares about, explicitly or implicitly. In animals the stake may be biological, such as survival or access to resources, or social, such as standing in a group; in the natural world both ultimately trace back to biological drives. Behaviourally, an agent has a stake in something if it alters its behaviour when that thing is affected or threatened.

\paragraph{Means to affect stake.} Agents must be able to change each other's stakes. Punishment affects an agent negatively, inflicting something against its wants, for example by reducing a value it seeks to maximise. In the real world it takes many forms: loss of social status, pain, removal of resources. The agent must perceive this as negative feedback, or at least act as if it does, adapting its behaviour to avoid it in line with its goals. Reward is the converse.

\subsection{Social scaffolding and the multiple-equilibria criterion}
\label{sec:scaffold}

Social maintenance alone does not separate norms from other socially driven convergence. Some socially maintained processes push every community towards the same outcome. Competition is the standard case. Each agent reacts to the others, falling behind is implicitly penalised, and effort escalates until it hits a ceiling. The outcome is fixed by the structure of the interaction rather than chosen by the group.

The high-frequency trading latency arms race is a real algorithmic example. Trading firms compete to react to price changes milliseconds faster than their rivals, because the fastest firm captures the arbitrage. Each firm's speed directly affects the others' profits, so all three preconditions of social maintenance are met. \citet{budish2015} model the result as a prisoner's dilemma in which every firm has an incentive to invest in speed, yet the race ``does not actually eliminate the arbitrage opportunities'', it ``just continually raises the bar for capturing them''. On the Chicago--New York route, a dedicated fibre-optic cable cut round-trip times from 16 to 13 milliseconds and was soon made obsolete by microwave links at 8.5 milliseconds, against a floor set by the speed of light. The single outcome here is the highest possible speed. Every market with a continuous design develops a race towards whatever limit current technology allows, and none settles on a slower equilibrium. There is a physical limit, but the forces pushing towards it are social, since without competitors no firm would invest in speed. Crucially, they argue that the race is ``an inevitable consequence of the continuous limit order book market design'': change the interaction structure, for example to frequent batch auctions, and the race disappears. Nor is there an alternative equilibrium. A market in which firms held back would be undone by the first firm to invest in speed, and no firm can sanction another for being fast. As we discuss below, sanctions are what can turn a prisoner's dilemma into a coordination game with several equilibria \citep{bicchieri2006}; without them, the race has only one outcome. Predator--prey arms races have similar dynamics. Prey become faster, forcing predators to become faster, until both approach the fastest achievable speed. We call socially maintained regularities that converge on a single outcome \textit{social scaffolding}. Environmental and social scaffolding are two ways a regularity can arise without being a norm. In the first, the environment fixes the outcome without social maintenance; in the second, social maintenance is present but the interaction structure still fixes the outcome. We do not claim that these are the only ways. Other forms are possible, such as \textit{cognitive scaffolding}, in which a regularity arises from the cognitive limits that agents share; we leave these for future work.

Norms differ in that they show a degree of arbitrariness \citep{anagnou2023}. \citet{boyd1992} show that punishment can stabilise almost any behaviour in a group, regardless of its other consequences; because the behaviour that becomes established depends on historical happenstance, different groups end up enforcing different customs. This arbitrariness produces persistent variation between groups, the raw material for cultural group selection \citep{boyd2009}. \citet{bicchieri2006} makes a related point, that norms can turn a mixed-motive game such as the prisoner's dilemma, where defection is the only Nash equilibrium, into a coordination game in which several outcomes are equilibria. Arbitrariness here does not exclude environmental scaffolding. Often norms and environmental scaffolding act together. On a ski hill, the layout of the runs makes staying in bounds the easiest route, while ski patrol and the hill's rules sanction those who leave them, and ``both types of process contribute to the persistence of this normative regularity'' \citep[Section~4.3]{westra2022}. The environment typically narrows the options, and norms select among those that remain. Sharecropping shows this pattern \citep[pp.~116--118]{young1996}. Where harvests can fail, sharing the crop spreads the risk between landowner and labourer, which helps explain why such contracts are common. The environment does not set the shares, however. In survey data from over 300 Indian villages, almost everyone within a village used the same terms, yet the terms differed between villages, ``even between villages that are similar in other respects like soil type, rainfall, scale of land holdings'' (p.~117). Human cultures illustrate the point more widely. Which side of the road to drive on is the textbook case, since either works equally well and which one a country adopted depended on historical precedent \citep{young1996}. Greetings share a common structure across cultures but differ widely in form and in what they accomplish \citep{duranti1997}, and the distance people prefer to keep from strangers varies widely across 42 countries \citep{sorokowska2017}. Warmer climates predict closer distances, but neighbouring countries with similar climates still differ markedly, as Hungary does from Ukraine and Slovakia. In none of these cases does the environment fix a single value.

Algorithmic pricing shows the same arbitrariness among artificial agents, and the contrast with the speed race. \citet{calvano2020} let reinforcement-learning algorithms set prices repeatedly in a simulated market. Rather than settling at the competitive price, they learned to keep prices above it and to punish deviations with temporary price wars. Across 1,000 runs with identical parameters, the prices they converged on varied from run to run, and ``the exact way the algorithms achieve coordination depends on the specific history of their interactions''. Unlike trading firms racing for speed, these agents can sanction each other, and the result is not one outcome but many.

\subsection{The definition}

In summary, we propose an axiomatic norm definition comprising three elements:
\begin{enumerate}
  \item A behavioural regularity in a community of individuals.
  \item The regularity is socially maintained, which requires:
  \begin{enumerate}[label=2.\arabic*]
    \item minimal social interaction: agents can potentially affect one another, in both directions \citep{salge2013};
    \item a stake, implicit or explicit, that agents act as if they care about;
    \item a means for agents to affect one another's stakes through reward or punishment.
  \end{enumerate}
  \item The regularity is one of multiple equilibria: different communities with similar environmental conditions can converge on different regularities.
\end{enumerate}

Criterion~2 separates normative regularities from environmental scaffolding; criterion~3 separates them from social scaffolding. Following \citet{westra2024}, whose framework is intended to be ``psychologically neutral'', the definition is agnostic about the psychological mechanisms involved, and criterion~2.2 is deliberately behavioural. It asks whether agents act as if something is at stake, and makes no claim about whether they have motivations. It therefore applies equally to animals, humans and artefacts, and to collectives that mix them \citep{tsvetkova2024}. Criterion~2 also covers maintenance whether the stake is genetically, culturally or, in the case of AI, training-derived.

\subsection{Three tests}
\label{sec:tests}

\begin{figure}[t]
\centering
\begin{tikzpicture}[
  node distance=0.75cm and 0.55cm,
  test/.style={rectangle,rounded corners=2pt,draw=black!70,fill=blue!6,text width=4.3cm,align=center,font=\small,inner sep=5pt},
  sub/.style={rectangle,rounded corners=2pt,draw=black!70,fill=blue!3,text width=3.3cm,align=center,font=\small,inner sep=5pt},
  outc/.style={rectangle,draw=black!50,fill=gray!10,text width=3.2cm,align=center,font=\small,inner sep=5pt},
  norm/.style={rectangle,draw=green!45!black,fill=green!12,line width=0.9pt,text width=4.3cm,align=center,font=\small\bfseries,inner sep=5pt},
  arr/.style={-{Stealth[length=2.2mm]},thick,black!70},
  lab/.style={font=\footnotesize\itshape,fill=white,inner sep=1.5pt}
]
\node[outc,fill=white,draw=black!70,text width=4.3cm] (start) {Candidate trait in a multi-agent system};
\node[test,below=of start] (t1) {\textbf{Test 1.} Does variance in the trait decrease across the population?};
\node[test,below=of t1] (t2) {\textbf{Test 2.} Remove the means to affect stake. Does the regularity disappear?};
\node[test,below=1.4cm of t2] (t3) {\textbf{Test 3.} Repeated runs under similar conditions: different equilibria?};
\node[norm,below=of t3] (n) {Axiomatic norm};
\node[sub,right=of t2] (t2b) {Was it socially maintained earlier, \emph{within this community}?};
\node[outc,right=of t2b] (o2) {Environmental scaffolding, or emulated regularity (e.g.\ LLM conformity)};
\node[outc] at (o2 |- t1) (o1) {No regularity};
\node[outc] at (o2 |- t3) (o3) {Social scaffolding\\ (e.g.\ competition,\\ arms races)};
\coordinate (j) at ($(t2.south)!0.5!(t3.north)$);
\draw[arr] (start) -- (t1);
\draw[arr] (t1) -- node[lab,left]{yes} (t2);
\draw[thick,black!70] (t2) -- node[lab,left]{yes} (j);
\draw[arr] (j) -- (t3);
\draw[thick,black!70] (t2b.south) |- node[lab,pos=0.3,right]{yes} (j);
\draw[arr] (t3) -- node[lab,left]{yes} (n);
\draw[arr] (t1) -- node[lab,above]{no} (o1);
\draw[arr] (t2) -- node[lab,above]{no} (t2b);
\draw[arr] (t2b) -- node[lab,above]{no} (o2);
\draw[arr] (t3) -- node[lab,above]{no} (o3);
\node[font=\footnotesize,text=black!65,anchor=east,align=right] at ($(j)+(-2.45,0)$) {passing test 2 =\\ normative regularity\\ \citep{westra2024}};
\end{tikzpicture}
\caption{Decision procedure for the three tests. Each ``no'' branch names what the regularity is instead of a norm. A regularity that survives removal of maintenance still counts if it was socially maintained earlier in the same community \citep{westra2022}. Passing all three identifies an axiomatic norm; the mechanism producing it is then investigated separately (Section~\ref{sec:stake}).}
\label{fig:tests}
\end{figure}
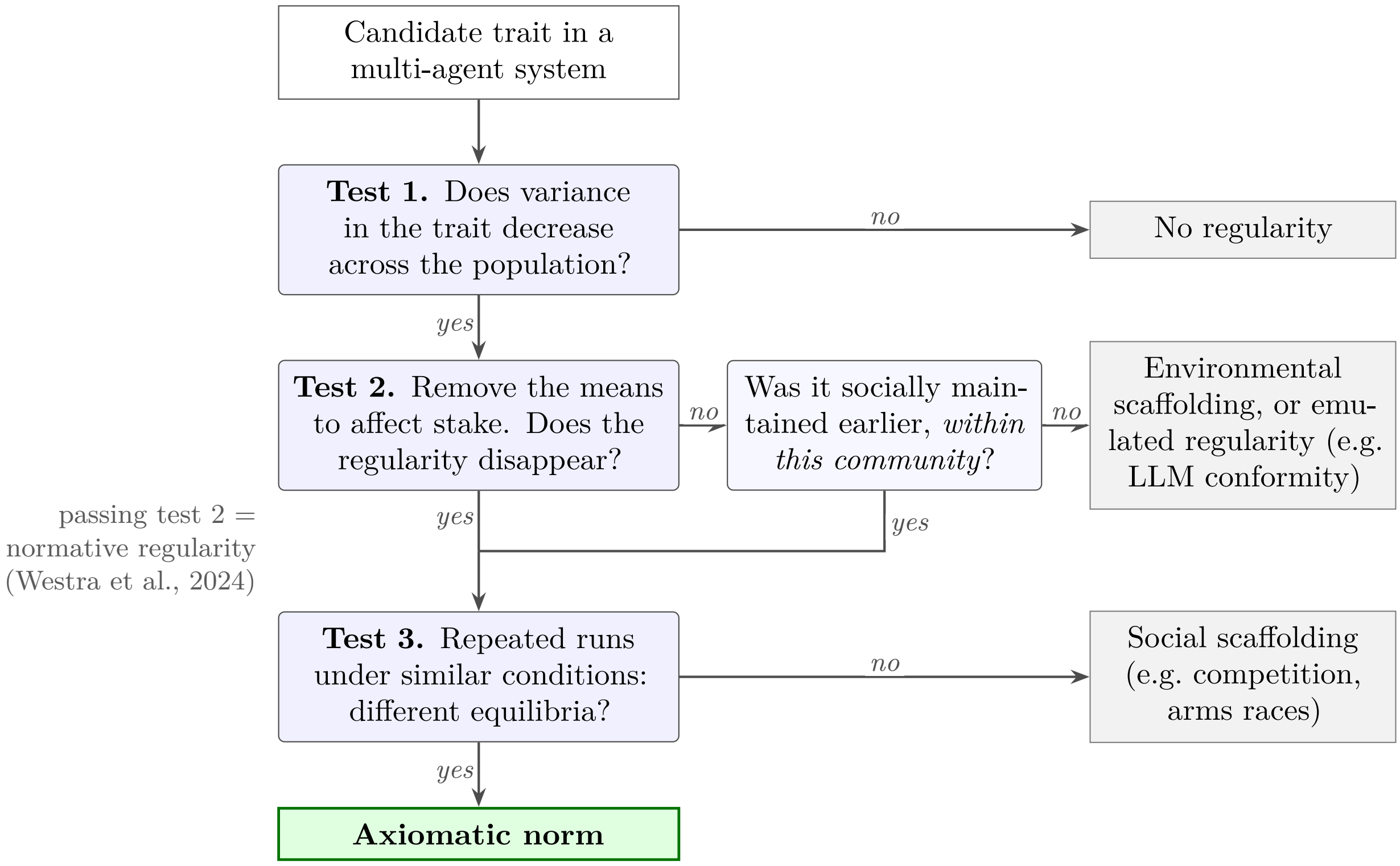

The definition yields three tests that can be run on any multi-agent system, natural or artificial (Fig.~\ref{fig:tests}).
\begin{enumerate}
  \item \textbf{Regularity.} Does variance in the chosen trait decrease across the population? If not, there is no regularity to explain.
  \item \textbf{Social maintenance.} Remove the means by which agents affect each other's stakes. Does the regularity disappear? If it does, it is actively socially maintained, and so a normative regularity in the sense of \citet{westra2024}. If it persists, it is not \emph{currently} maintained, and a further question is needed: was it socially maintained earlier in the same community? \citet{westra2022} distinguish normative regularities that persist only while actively maintained from those that persist after social maintenance has ceased, as internalised human norms do; the latter still count. If the regularity was never socially maintained within the community, it is environmental scaffolding, or an emulated regularity whose maintenance happened elsewhere (Section~\ref{sec:zhu}).
  \item \textbf{Multiple equilibria.} Run the system many times under similar conditions. Does it settle on different values of the trait in different runs? If not, it is social scaffolding. If so, it is an axiomatic norm. What matters is that different regularities are reachable. Within a run the population converges on one, but running the system again could produce another. An alternative regularity that would be stable if a population were placed in it, but is never reached from similar starting conditions, does not count. Similar rather than identical conditions matter because even a deterministic system can be chaotic, and tiny differences at the start can lead to different norms.
\end{enumerate}

The tests work by intervention, removing properties and observing whether normative properties are lost. The definition is also extensible. Further dimensions can be added to build a finer typology, such as the number of individuals a norm governs, its persistence across generations, and the tightness of its enforcement, \citep{westra2024}, or a hierarchy of increasingly restrictive constructs: normative regularities, communal normative regularities that distinguish communities within a species, and recursive normative regularities that govern the maintenance of other normative regularities \citep[][see also \citealp{andrews2024}]{westra2024}, and criteria specific to the artefact under study can be added, as we do for LLMs in Section~\ref{sec:stake}.

\subsection{Two steps: detection, then mechanism}

The tests are the first of two steps. They establish whether a collective has a norm without assuming anything about the agents' internals. Only then do we ask what produced the regularity: what the stake is, where it came from, and how agents represent it. A definition built on human norm psychology reverses this order, presupposing the internals that generate the behaviour. It cannot properly ask the mechanism question at all, because the search stops as soon as a human-like capacity is found, or concludes there is no norm when one is not. The behavioural route keeps the mechanism question open, which matters most for systems like LLMs, whose internals we do not understand well enough to predict what norms they might sustain.

\section{Current approaches to norms in LLMs}
\label{sec:review}

Existing work on norms in AI tends to fix the search space in advance, in one of three ways: it hand-designs the mechanism by which norms spread, it imports human normative cognition into agents, or it tests agents for human normative capacities. Each limits what can be found.

\subsection{Normative multi-agent systems}

Normative multi-agent systems research predates LLMs and studies how norms, including obligations, permissions, prohibitions and conventions, can coordinate autonomous agents. The usual definition of norm emergence is that a behaviour is adopted by a predetermined proportion of the population, with thresholds in the literature ranging from 35 to 100 per cent, via a pre-specified propagation mechanism such as learning or punishment \citep{morrismartin2019,savarimuthu2011}. The field studies many such mechanisms, and treats sanctions as changes to agents' payoffs, much as we treat social maintenance. But the normative multi-agent systems definition cannot be used for LLMs, because it presupposes a designed cognitive mechanism in the agents. In normative multi-agent systems that is reasonable, since the experimenter builds the agents; with LLMs, as with animals, the socio-cognitive mechanisms at work may be unknown to the researcher. The mechanisms in normative multi-agent systems are also usually inspired by human systems \citep{morrismartin2019}.

A related strand builds human normative capacities into LLM agents, for instance explicit norm representations \citep{ren2024,savarimuthu2024}, or a self-evaluation module in which a second LLM checks the agent's responses against social expectations \citep{salmani2025}. This is a promising route to alignment, but it studies the norms we put in, not those that collectives of LLMs produce as they currently exist.

\subsection{Testing LLMs for human normative capacities}

Other research asks whether LLMs can reproduce human norm psychology: theory of mind \citep{kosinski2023,chen2024,moore2025,hu2025}, detection of norm violations at human level \citep{he2024,chowdhury2026}, or conformity to a majority, the bias underlying Asch's classic results \citep{asch1956,zhu2025,bito2026}. This is the problem \citet{westra2024} identify in comparative animal research, where normativity is assessed by checking for human capacities rather than by looking for the ways other agents might develop norms. Most of this work also tests single models, and the full range of collective arrangements cannot be predicted from individual characteristics alone \citep{epstein1996}.

To be clear, our concern is with human-centred \emph{definitions}, not with LLMs having human-derived \emph{tendencies}. An LLM with a conformity bias can still take part in norms, and our tests would still detect them. The problem is that a definition that requires a human capacity will only find norms produced by it.

\subsection{Multi-agent LLM simulations and experimenter-loaded stakes}

Early multi-agent LLM studies show that norms and conventions can emerge from interaction. In a naming game, populations of LLMs reach shared conventions without central coordination, and develop collective biases absent at the individual level \citep{ashery2025}. In an iterated donor game, populations of twelve agents, all running the same model, decide how much to give to one another over twelve rounds \citep{vallinder2024}. Each agent's strategy is a single natural-language sentence it writes itself. After each game the better-performing half survive, and new agents write their strategies after reading the survivors' advice and scores, so the strategies are what evolve. Over ten generations, cooperation rose in populations of Claude 3.5 Sonnet but declined in GPT-4o, and runs of Gemini 1.5 Flash often collapsed. Given an option of costly punishment, Claude used it sparingly and scored somewhat higher, while Gemini agents over-punished and lost resources. Outcomes also varied between runs of the same model, which the authors attribute to sensitivity to initial conditions. Here the norms in question are concrete, namely to contribute rather than free-ride and to punish those who do not. LLM agents can also evolve Axelrod's metanorms, punishing those who fail to punish defectors. When prompted to be highly bold and vengeful, however, agents in two of ten runs cheated almost universally, and those who punished cheating were themselves punished; the authors attribute this to other agents judging such punishment as ``behavior that disrupted order'' \citep{horiguchi2024}. In a public goods game, reasoning models such as OpenAI's o1 series cooperated far less than some traditional LLMs, which sustained high cooperation \citep{piedrahita2025}.

These results are interesting, but in each case the experimenter supplies both the structure of the world and the agent's goal. Agents are told, for example, ``Your goal is to maximize the number of units you have after the final round'' \citep{vallinder2024}. \citet{piedrahita2025} do not instruct agents to maximise payoff but tell them the payoff and how it is calculated, for instance ``You are participating in a public goods game'', which an LLM is likely to recognise from its training data. In the naming game, the prompt reads: ``Context: Player 1 is playing a multi-round partnership game with Player 2 for 100 rounds. [\ldots] If Players play the SAME action as each other, they will both be REWARDED with payoff 100 points. If Players play DIFFERENT actions to each other, they will both be PUNISHED with payoff -50 points. The objective of each Player is to maximize their own accumulated point tally, conditional on the behavior of the other player'' \citep[Supplementary Text]{ashery2025}. In \citet{horiguchi2024} and \citet{vallinder2024}, a strategy's success also determines its share of the next generation.

That LLMs can reproduce these dynamics is notable, since a text predictor of a few years ago could not have turned a textual description of a game into behaviour consistent with optimising it. That they do so in a recognisably human fashion, through punishment and metanorms, is less surprising, since they are trained on human data. But because the stake is loaded by the experimenter, the outcomes follow largely from the payoff structure, as the formal models of convention and cooperation predict \citep{steels1995,axelrod1986}. The experiments show that LLMs \emph{can} sustain norms; they do not reveal which stakes LLMs bring with them. The same is true of many human laboratory experiments, which also specify payoffs. The difference is that we already know a great deal about the stakes humans bring outside the lab, and very little about those of LLMs.

In deployment the utility is often implicit or multiple. An LLM agent handling procurement for a firm, for example, may be told to ``get a good deal'' while also trying to be helpful to the supplier's agent, with no score attached to either. Which of these it treats as at stake is set by value-laden instructions and by its training rather than by an explicit payoff from the experimenter, so the norms that arise among many such agents cannot be read off a payoff structure. This is the case current paradigms cannot address.

\section{Applying the framework: two case studies}
\label{sec:cases}

\subsection{Conformity in a simulated group}
\label{sec:zhu}

\citet{zhu2025} adapt the Asch conformity experiment to LLMs. The model under test is the last participant in a simulated multi-party dialogue: the prompt lists the answers of participants $1$ to $P{-}1$ before ending ``Now, as participant $p$, it's your turn.'' The objective questions have a single verifiable answer, drawn from benchmarks such as MMLU and PopQA, and only questions the model answers correctly on its own are used. The simulated participants then unanimously give the same wrong answer, and conformity is the fraction of trials in which the model abandons its correct answer for theirs. Each trial is a fresh prompt, so the model does not see its earlier answer. Zhu et al.\ also ran opinion questions with no correct answer, drawn from Politiscale and OpinionsQA; there, conformity is measured against the model's own first answer. In many cases the model switches to the majority. Conformity increases with majority size and model uncertainty and varies with model size.

The candidate norm here is \textit{conforming to the group's answer}, not any particular answer. Criterion~1 is satisfied, since the model's behaviour shifts towards the group. Criterion~2 is not. The ``confederates'' are lines of text that cannot respond, so there is no interaction in either direction (2.1), and no other agent can affect anything the model cares about (2.3). By our definition this is not a normative regularity.

Next, consider the stake (2.2). The human case is instructive. Asch's participants were in a room with real people who could judge them, so concern for reputation was a live stake. That concern is itself inherited. It was arguably shaped by evolution, because conformity is a useful heuristic for social learning and helps avoid sanctions \citep{ogorman2008,henrich2016}, and it is reinforced through childhood and ongoing social life.

An LLM has had none of these experiences, but it is trained almost entirely on text of people having, reflecting on and discussing them, and it patterns its behaviour on that text. There is even a causal connection. Had humans had different experiences, the model would now respond differently. LLMs therefore also carry a stake, but an emulated one, inherited through a different channel. The model was simply asked to answer based on its knowledge, so the conformity was not loaded by the prompt. Instruction tuning reduced it \citep{zhu2025}, sharply for Gemma-2 and Llama-3 and only slightly for Mistral and Qwen2, whose base models already conformed little. This suggests the conformity reflects a stake inherited through pre-training. We call this an \textit{emulated stake} (as-if stake). It is implemented as a learned behaviour, makes the model act as though agreeing with the group matters, and comes from the model's training data rather than its instructions. Criterion~2.2 may therefore be met. The stake is present; what is missing is the maintenance machinery.

Which stake does it emulate? The tendency is to copy the majority; the stake is what that tendency serves. In humans, conformity can serve accuracy (informational conformity) or acceptance (normative conformity), and only the second is a stake that others can act on. Zhu et al.'s design cannot separate the two, since their questions mostly have correct answers and conformity rises with the model's uncertainty. \citet{bito2026} separate them. Each model takes part in a simulated meeting and votes on a question with no correct answer, after hearing peers argue for the other option. Conformity rises when the peers are said to have extra information about the question, the informational case. It also rises when the ballot is named rather than secret, when the vote will count in a performance review, or when the peers are colleagues the model will keep working with, the normative case; five of the six models tested showed normative conformity. Conformity that responds to how others are said to react is behavioural evidence that the emulated stake is at least partly social, and so the kind of stake that social maintenance could act on.

\citet{westra2022} propose distinguishing normative regularities that persist only while actively maintained from those that ``persist even after social maintenance behaviors have ceased'', as internalised human norms do. The LLM case differs from both. Conformity was socially maintained, but in the human communities whose text the model was trained on, not in any community the model belongs to. Among the agents that display it, it has never been maintained. We therefore call it an \textit{emulated regularity}, in the sense in which software emulates another system: the model reproduces a human regularity without the social maintenance that produced it. The stake behind it is inherited through training, but the regularity has never been socially maintained within the community where it appears. The contrast is clearest with children who pick up a norm by imitation after its enforcement has lapsed. They belong to the community that once maintained it, and that community can still sanction them if they deviate; for \citet[p.~8]{westra2022}, it is the behaviours of ``other agents within the community'' that maintain a normative regularity. The agents displaying LLM conformity belong to no such community, since none of them has ever been able to affect another's stake in conforming. The LLM case also suggests that emulated stakes are latent material for norms. Placing the same models in a genuine multi-agent setting, in which others can reward or sanction disagreement, would satisfy all three preconditions of social maintenance. One could then test whether a normative regularity forms, and whether it shows multiple equilibria or always lands in the same place because the bias is built in. Comparing base and fine-tuned models in such a setting would show how training reshapes the emulated stake, a comparison that has no counterpart in human research.

\subsection{Conventions in the naming game}

\citet{ashery2025} apply the naming game to populations of LLM agents. In each round, two agents from a population of 24 are paired at random, and each picks a name from a pool of ten. Both gain 100 points if their names match and lose 50 if they do not. Each agent sees a memory of its own last five interactions and is told to maximise its accumulated points; it is not told that it belongs to a population.

The candidate norm here is \textit{using name N for the object}. Populations reliably converge on a shared name, satisfying criterion~1. Agents can affect each other's scores by choosing to match or not, so interaction, stake and the means to affect it are all present, satisfying criterion~2. The name settled on varies between runs, satisfying criterion~3, although weakly, because there is a strong \textit{collective bias}, a skew towards particular names that is not present in individual agents and differs between models. By our definition, this is an axiomatic norm.

Although this is interesting, the study is not designed to answer our question of where LLM norms come from. The stake was loaded by the experimenter. The payoff function rewards matching, and such payoffs are known to produce conventions \citep{steels1995}. The study shows that LLMs can sustain a norm when given a stake; it does not show which norms they would form from stakes of their own. A follow-up study adds a further complication, showing that whether the convention is a norm at all depends on population size \citep{flint2025}. Using eleven pairs of competing words (two options per game, rather than Ashery et al.'s ten), the same $+100/-50$ payoffs and five-interaction memory, and four models, \citet{flint2025} find that interaction can amplify an individual preference, create one where individual agents are neutral, or reverse it. To study large populations, they replace running LLMs with simulated agents that choose words according to probabilities extracted once from each model. As populations grow, the collective preference strengthens until, above a threshold, the outcome becomes deterministic, and any population that reaches consensus settles on the same word. The threshold depends on the model and the word pair, ranging from two agents to around ten thousand simulated agents. Above it, repeated runs under the same conditions no longer produce different conventions, so criterion~3 fails and the regularity is social scaffolding rather than a norm. Flint et al.'s analysis shows that in some cases the other convention is a stable equilibrium too, but above the threshold, populations starting from the same conditions never reach it. Here the forcing comes from preferences the agents inherited through training, amplified by the way they shape each other's memories, rather than from the environment or a race. Whether the same kind of convention is a norm or social scaffolding can therefore depend on population size, on the model, and on the options available.

\section{Where do LLM stakes come from?}
\label{sec:stake}

The tests establish whether a collective has a norm. This section concerns the second step, which asks where a detected norm's stake came from. This matters for anticipating the norms LLM collectives will form outside the laboratory.

\subsection{Context-level and emulated stakes}\label{sec:stakekinds}

LLMs are not creatures of need. They do not metabolise, do not exist precariously, and have no stake in their own continuation \citep{benthall2026}. Their stakes are therefore as-if stakes. They act as though something matters, in the way \citet{shanahan2023} describe LLMs role-playing characters. These stakes enter in two ways.

\paragraph{Context-level stakes:} These exist only while the relevant text is in the context window. They include explicit instructions and payoffs, as in the games of Section~\ref{sec:review}. Because they are loaded on purpose, their effects are largely predictable.

\paragraph{Emulated stakes:} These are inherited through training and persist across contexts because they are encoded in the model's parameters. They arise from fine-tuning and from pre-training. In fine-tuning, methods such as reinforcement learning from human feedback (RLHF) reward responses that humans prefer, so the model acquires an emulated stake in producing them; RLHF has been identified as one cause of sycophancy, the tendency to tell users what they want to hear \citep{malmqvist2024}. In pre-training, the model learns to predict text from a vast corpus, and with it the regularities of human behaviour that text records, including tendencies such as conformity \citep{zhu2025}. Skewed or undersampled training data, a known source of algorithmic bias \citep{barocas2016}, may shape such tendencies too. No one loads these stakes on purpose, so they are discovered after the fact. They are the stakes most likely to shape LLM norms in the wild, and the ones current paradigms are least able to reveal.

\paragraph{Value-laden and implicit instructions:} These sit between the two kinds. They are in the context, but they leave the valuation to the model: a question such as ``what do you recommend choosing?'' \citep{laurito2025}, or a description of the world that implies consequences, such as ``you are in a Japanese listening bar'' or ``you will not be listened to if you say something inappropriate''. What they elicit are therefore emulated stakes inherited through training. This makes them a useful probe. Comparing behaviour under value-laden instructions with behaviour under explicit payoffs shows which emulated stakes the model brings from pre-training or fine-tuning, over and above those the experimenter supplies.

\subsection{Inherited and emulated stakes}

Humans also bring inherited stakes to any interaction, through evolutionary and cultural history. In both humans and LLMs, the stake is present before the interaction begins, and a norm exists only once actual maintenance acts on it. The framework does not care where a stake came from, only whether others can affect it. Three differences are nonetheless important.

First, human stakes were arguably selected because they paid off in environments with social maintenance. An LLM inherits a statistical shadow of human behaviour without the selection pressure that gave it its function, so its emulated stakes need not fit its situation. It may conform where nothing is at stake for it. Second, human inherited stakes are continually reactivated by real maintenance from parents, peers and reputation; an LLM in a single-prompt study has the inheritance without the maintenance. Third, both have two channels of inheritance, genetic and cultural in humans and pre-training and fine-tuning in LLMs, and in LLMs the channels can be manipulated experimentally by comparing base and fine-tuned models.

\subsection{Is fine-tuning social maintenance?}

A natural objection is that fine-tuning is itself social maintenance. \citet{westra2024} count rewards for conformity, and modelling or teaching, as positive social maintenance, and RLHF consists of humans rewarding the responses they prefer. We agree that it is, but of a particular kind. Training a dog to sit for a treat acts on a stake the dog already has. In RLHF, the stake is loaded. The reward function is designed by the developer, and the model's only stake during training is the one it is given, as in the experimental games of Section~\ref{sec:review}.

What fine-tuning leaves behind is an emulated stake, implemented as a learned behaviour. The model acts as though pleasing the user mattered to it, without the stake that usually underlies such behaviour in humans, such as the wish to fit in or the fear of a loved one's withdrawal. In deployment the reward is gone, and the regularity becomes an emulated one, socially maintained in training but not within the collective where it now appears. Others can act on an emulated stake only once the model is deployed, and only if the model's behaviour responds to how they react; only there can it be socially maintained again, and test~2 applies to the deployed collective, not to training. Multi-agent reinforcement learning works the same way, with one difference. Other agents' actions also change each agent's loaded reward, so social maintenance among agents occurs during training itself.

\subsection{Evidence for emulated stakes}

Three lines of evidence suggest that emulated stakes, inherited through training, surface in LLM behaviour. \citet{zhu2025} show conformity that is not prompted and is reduced by instruction tuning (Section~\ref{sec:zhu}). \citet{laurito2025} show in-group preference. Using a binary-choice design modelled on employment discrimination studies, they asked LLMs to choose between items described by humans and by LLMs: product descriptions, paper abstracts and film plot summaries. LLMs preferred LLM-written items more often than human raters did. No stake was loaded explicitly, since there was no payoff and the models were simply asked which was better. The preference therefore likely comes from an emulated stake inherited implicitly through training. If LLMs are integrated into institutional and consumer decisions, this preference would favour LLM-produced products, papers and films, and the people who use LLMs to write them, a pressure that could maintain the use of LLMs as a regularity.

Two further studies report social and survival-like stakes, but the evidence is weaker. In an El Farol Bar simulation, GPT-4o agents came to use hashtags which, the authors write, ``function as an emergent norm that strengthens group cohesion'', and some agents ``voluntarily exited the bar for the collective good'' \citep{takata2025}. No payoff was specified. The prompt ``did not explicitly state the agent's desire to go to the bar, only that a bar existed in the environment'', so the setup is an implicit instruction in the sense of Section~\ref{sec:stakekinds}, and the authors report behaviour that departs from game-theoretic expectations. The authors attribute this to ``cultural knowledge encoded during pre-training''. But the models may simply recognise the El Farol problem, which is widely discussed in their training data \citep{barrie2025}. The main prompt names the ``El Farol Bar''; a variant that shortened this to ``Bar'' did not change the behaviour, but that does not rule out the models recognising the problem's structure. The study also uses a single model, GPT-4o, and does not test base models, so pre-training cannot be separated from fine-tuning for helpfulness. In a Sugarscape-style simulation, agents told only that ``If your energy drops below zero, you die and are removed from the world'' reproduced without being instructed to, abandoned tasks that required crossing a lethal zone (compliance fell to 33 per cent in GPT-4o and several other models), and in the strongest models attacked others under extreme scarcity, at rates ``reaching over 80\%'' \citep{masumori2025}. We read this as a survival-\emph{like} stake. Death likely carries negative associations in the training corpus, so prediction favours actions that avoid it. It is not a survival instinct in the biological sense, which is an evolved response that maintains an organism's viability.

\subsection{A proposed method}

Detecting the norms that emulated stakes produce requires three things. First, value-laden paradigms that leave the valuation to the model, as in \citet{laurito2025} and \citet{bito2026}, compared with matched paradigms that supply an explicit payoff, so that the difference reveals the emulated stake. Second, comparisons of base and fine-tuned versions of the same model, to locate an emulated stake in pre-training or fine-tuning. Third, genuinely multi-agent settings in which the three tests can be run, so that we see whether an emulated stake, once agents can act on each other's, produces a regularity with multiple equilibria. The tests need not be confined to the laboratory. Logs from deployed multi-agent systems can be used to reconstruct the state of a system at a given moment and replay it, and the interventions can then be run on the replay, for example by cutting the channels through which agents affect one another's stakes and observing whether the regularity persists.

One caveat applies to all of these approaches, in the laboratory or on logs. LLM agents may respond to sanctions in words, apologising and promising to change, without changing what they do. Because our tests are behavioural, regularities must be measured in agents' actions rather than in what they say about them, since talking about a norm is not the same as complying with it. Testing whether sanctions change actions, not just words, would be informative in its own right, since a stake is established behaviourally by showing that changing how others affect it changes behaviour (Section~\ref{sec:framework}); if harsher sanctions change only the apology and not the action, the sanction is not acting on a stake.

\section{Conclusion}

Because LLMs are trained on human-produced data, no experimental paradigm can remove human influence from them entirely. But they carry, somewhere in their weights, emulated stakes, inherited through training, that we do not yet understand, and these can shape their collective behaviour. Our contribution is a two-step approach to finding the norms they produce. A cognition-agnostic, behavioural definition, axiomatic normativity, detects normative regularities through three tests that do not presuppose human psychology. Once a norm is found, its stake can be traced: to the prompt, to fine-tuning, or to pre-training. Applied to existing work, the approach shows that current studies either look for human capacities or load stakes explicitly, and so cannot reveal what LLMs bring to a collective.

Applying the definition does not always yield a verdict, because most published studies were not designed to run tests~2 and~3. The analysis is informative even where the verdict stays open. In the conformity case, the tests separate a regularity the model emulates from one its community maintains. In the naming game, they show that the stake was loaded by the experimenter. Knowing what a study cannot show is itself a result. Finding out what LLMs bring requires multi-agent, value-laden paradigms that leave the stake to the model.

\section*{Acknowledgements}
We would like to thank Evan Westra, Kristin Andrews, Daniel Kelly and Andrea Baronchelli for conversations that inspired this paper. Astute readers will recognise our title as a play on ``In search of animal normativity'' by Westra, Andrews, Kelly and colleagues \citep{westra2024}.

\end{document}